\documentclass[conference]{IEEEtran}

\usepackage{cite}
\usepackage{amsmath,amssymb,amsfonts}
\usepackage{algorithmic}
\usepackage{graphicx}
\usepackage{textcomp}
\usepackage{xcolor}
\usepackage{float}

\def\BibTeX{{\rm B\kern-.05em{\sc i\kern-.025em b}\kern-.08em
    T\kern-.1667em\lower.7ex\hbox{E}\kern-.125emX}}
\begin{document}
\bstctlcite{bstctl}
\title{{EdgeVLN: Runtime-Aware Deployment Ready Quantized Vision Language Navigation Model}}

\author{
Rithvik Jonna$^{1,*}$,
Man Namgung$^{2,*}$,
Aakash Gurram$^{2}$,
and Tinoosh Mohsenin$^{1,2}$\\
$^{1}$Department of Electrical and Computer Engineering \quad
$^{2}$Laboratory for Computational Sensing and Robotics\\
Johns Hopkins University\\
$^{*}$Equal contribution
}

\maketitle

\begin{abstract}
Vision-language navigation (VLN) models perform well but target compute-rich
platforms, limiting deployment on memory- and power-constrained robotic edge
devices. Compression alone does not establish whether a VLN model fits an edge
platform's memory, latency, and energy budgets while preserving navigation
behavior. We introduce EdgeVLN, a runtime-aware, deployment-ready quantized
VLN model that closes this gap. EdgeVLN combines a quantized StreamVLN model
with Latent Trajectory Termination Extractor (LATTE), a lightweight causal
transformer that improves real-time stopping by predicting a Stop Action
verifier rank. Both execute through our \texttt{llama.cpp} VLN driver, which
reconstructs streaming context and prunes memory tokens on-board. We
characterize a pretrained StreamVLN backbone across weight quantization from
$8$ to $2$ bits and multiple inference runtimes to identify a feasible
operating point. LATTE reuses backbone hidden states within the budget freed by
quantization, requiring neither a second vision encoder nor an additional
backbone forward pass. We evaluate six backbone precisions and seven candidate
stop heads on BF16 and IQ4\_NL across all $1{,}839$ R2R VLN-CE val-unseen
episodes, where agents follow language instructions through photorealistic
indoor scans. We measure success rate~(SR) in simulation and latency, energy,
and resident memory on an NVIDIA Jetson Orin NX 16\,GB. LATTE achieves our
highest SR, $58.02\%$ on the deployed four-bit model, exceeding the BF16
baseline with only $0.013$\,s additional latency per navigation step.
Four-bit formats achieve nearly identical SR, but step energy varies
$36.8\times$ by execution path. Only IQ4\_NL under our VLN driver fits the
board: $11.35$\,GB resident, $20.8\times$ faster, and $13.3\times$ less energy
than storage-streamed BF16. INT2 collapses. Runtime selection, memory-token
pruning, and quantization are essential for efficient edge deployment.
\end{abstract}



\section{Introduction}

Language conditioned mobile robots increasingly rely on vision-language navigation (VLN) models to translate visual observations and natural-language instructions into navigation actions. However, many openly available VLN models are designed without the resource constraints of the platforms on which they are deployed. Current systems commonly combine a SigLIP class vision encoder with a dense language decoder of several billion parameters; the model we study needs $16.1$\,GB for weights alone in BF16 \cite{streamvln}. This exceeds the practical memory budget of representative edge platforms such as the NVIDIA Jetson Orin NX 16\,GB, whose shared LPDDR5 memory must support the GPU, CPU, operating system, camera pipeline, key--value cache, and compute buffers. As a result, navigation inference is often moved to a remote server, with the robot continuously streaming RGB-D observations over a network connection. This architecture introduces communication dependent latency, requires persistent connectivity, and transfers the robot's visual observations off the platform. 

Weight quantization is a natural way to reduce the memory footprint of VLN models, but embodied deployment is not determined by bit width alone. On this board, PyTorch doesn't provide a 4-bit tensor-core path, so the benefit of 4-bit weights is primarily lower storage and memory traffic, while the framework itself consumes roughly $3$\,GB of the $16$\,GB budget. Nominal weight size also fails to predict peak residency, because runtime overhead, KV-cache allocation, and compute buffers decide whether a model fits. Compression is further limited because only selected layers are quantized, while the attention projections, embeddings, LM head, vision tower, and projector remain at higher precision. Finally, quantization must be judged in closed-loop navigation, where a single incorrect action changes future observations and can compound over the trajectory.

We characterize StreamVLN \cite{streamvln} under INT8, INT4, NF4, IQ4\_NL, and INT2 weight quantization evaluating on step-by-step navigation on R2R VLN-CE Dataset val-unseen split \cite{vlnce} on habitat simulator, measuring success, latency, energy, and resident memory on a Jetson Orin NX 16GB. Treating the inference runtime as a design variable alongside bit width, a hardware--software co-design view of edge deployment, we identify IQ4\_NL with \texttt{llama.cpp} \cite{llamacpp} as a practical operating point that preserves near-BF16 navigation performance while sharply reducing on-board latency and energy. The same characterization also isolates the remaining accuracy headroom. Across viable precisions, the gap between success rate (SR) and oracle success rate (OSR) remains nearly unchanged, from $7.35$ points at BF16 to $6.80$ points at deployed IQ4\_NL, indicating that incorrect termination is a property of the base model rather than a quantization artifact; only at 2 bits, where the model collapses, does the gap increase. Motivated by this, we introduce LATTE (Latent Trajectory Termination Extractor), a lightweight real-time Stop Action verifier for continuous VLN-CE that rescores termination decisions from activations already produced by the backbone, operating within the runtime and memory budget reclaimed by quantization.

\vspace{0.5em}
The main contributions of this study are as follows:
\begin{itemize}
    \item We propose an end-to-end deployment pipeline that runs an $\mathbf{8.03}$B VLN model \cite{streamvln} on an Orin NX 16GB and evaluates five weight quantization formats across two inference runtimes under full closed-loop rollout on all 1,839 R2R VLN-CE Dataset val-unseen split  episodes, where an agent follows language instructions through photorealistic indoor environments to reach a target location, with latency, energy, and resident memory measured on-board.

    \item We introduce Latent Trajectory Termination Extractor(LATTE), an inline real-time Stop Action verifier that operates without an additional backbone forward pass. On our deployed EdgeVLN system, LATTE improves SR to $\mathbf{58.02\%}$ ($\Delta\mathrm{SR}=\mathbf{+0.71}$ points) for $\mathbf{0.013}$\,s more per navigation step.

    \item Our EdgeVLN model provides a deployable quantization--runtime operating point, IQ4\_NL under our \texttt{llama.cpp} VLN driver, which reconstructs the streaming context and prunes memory tokens on-board, that reduces resident memory by $\mathbf{1.32\times}$, latency by $\mathbf{20.8\times}$, and energy per navigation step by $\mathbf{13.3\times}$ relative to BF16.

\end{itemize}

\section{Related Work}

This section reviews three lines of work: efficient VLA/VLN deployment, vision--language model quantization, and oracle-gap reduction in navigation. These address complementary aspects of embodied inference, but not the joint problem of deploying large VLN models on resource-constrained robots.



\subsection{Efficient VLA and VLN at the Edge}

VLN Models~\cite{streamvln,navila} couple a video--language backbone with a bounded temporal context to produce
low latency navigation decisions, but their language models are sized for server class GPUs rather than embedded platforms. 
Efficient vision--language--action (VLA) work has focused on manipulation: \cite{litevlaedge,spvla,dyqvla} apply quantization, architectural simplification, or early exit to tabletop and arm control, and Vishwanathan et al.\ attribute roughly $75\%$ of step latency on Jetson to memory bound autoregressive generation \cite{vishwanathan2026characterizing}.
EdgeNav-QE \cite{edgenavqe} combines QLoRA
quantization with dynamic early exit inference for large navigation models and
evaluates navigation in Habitat-Sim. However, these studies report latency
under a single precision and runtime, and none characterize energy per
decision across multiple weight formats and inference runtimes on a
memory-constrained platform. 

\subsection{Quantization for Vision--Language Models}

Post-training methods such as
Activation-aware Weight Quantization (AWQ)~\cite{awq} and low-bit inference
formats such as GGUF/Q4\_K\_M primarily target the language backbone while
typically retaining higher precision in the vision encoder. Recent
hardware aware studies also show that quantization does not translate
directly from nominal model compression to inference efficiency, as
dequantization and hardware--software interactions can affect runtime
performance~\cite{shin2026rethinking}. 
This motivates selecting quantization and runtime jointly.


\subsection{Oracle Gap and Stop Action Verification Methods}

Mind the Gap~\cite{mindthegap} demonstrates that an auxiliary verification
module can reduce the gap between Success Rate (SR) and Oracle Success Rate
(OSR) in discrete panoramic navigation by reevaluating previously visited
viewpoints and selecting a better candidate. 
Edge VLM studies characterizing quantization use static image--text evaluation ~\cite{shin2026rethinking} rather than closed-loop rollout.
Moreover, the viewpoint selection mechanism of Mind the Gap does not directly transfer to continuous
VLN-CE, where revisiting an earlier viewpoint requires physical navigation.
We therefore adapt the verification idea to continuous VLN-CE by predicting
the Stop Action from the backbone's temporal decision-step embeddings, without
revisiting or reselecting previous viewpoints, and design it for inline
operation on an edge deployed VLN system.


\section{Methodology}

\begin{figure}[!b]
  \centering
  \includegraphics[width=1\linewidth]{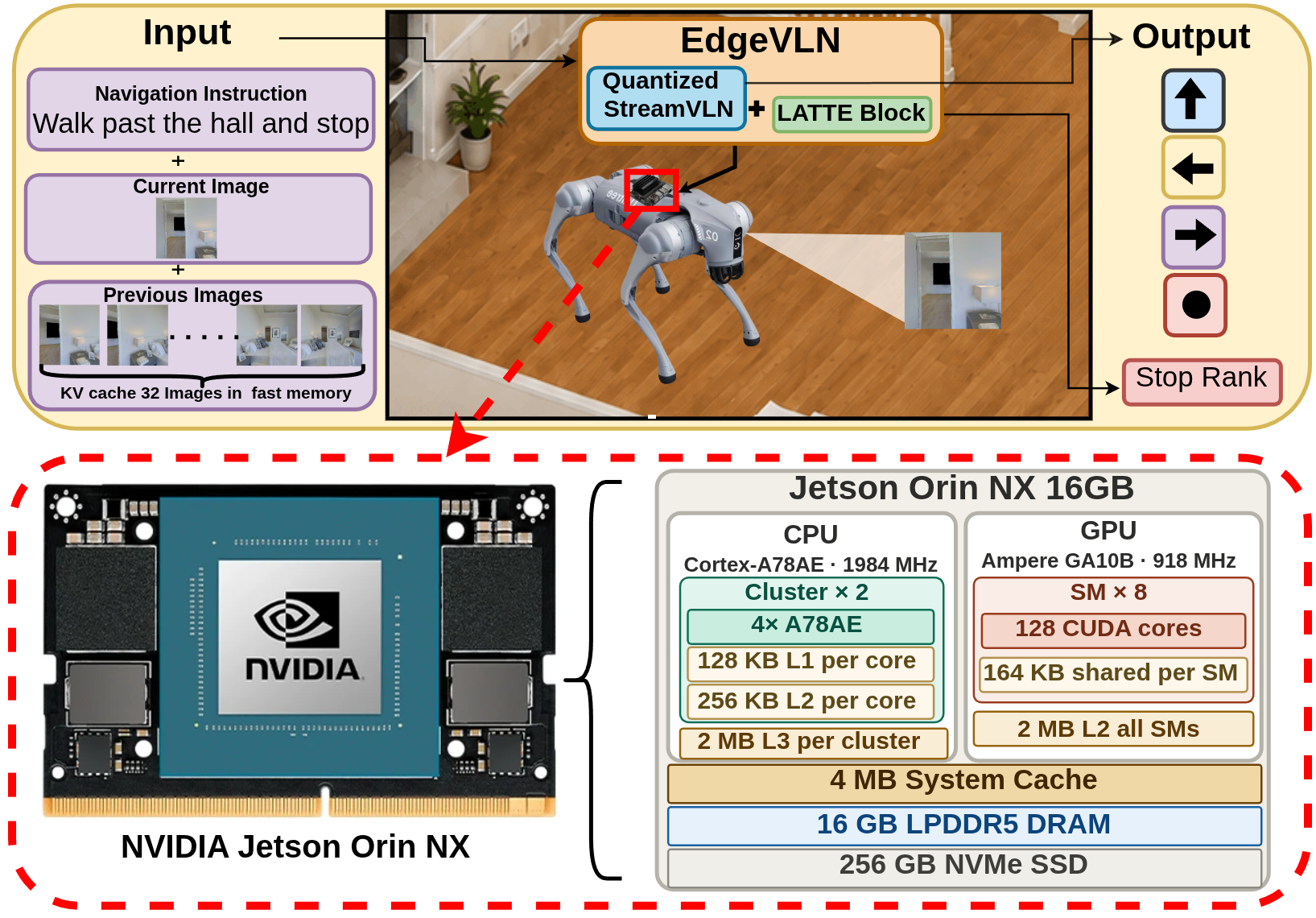}
  \caption{Overview of EdgeVLN. EdgeVLN consists of a quantized StreamVLN model and LATTE, an inline Stop Action verifier, both executed by our \texttt{llama.cpp} VLN driver on the NVIDIA Jetson Orin NX 16\,GB detailed below. The input is a natural-language navigation instruction, the current RGB frame, and a $32$-frame KV cache of earlier frames; the output is a navigation action---move forward, turn left, turn right, or stop---together with LATTE's Stop Action verifier rank.}
  \label{fig:deployment}
  \vspace{-1.5em}
\end{figure}

In this section, we build EdgeVLN in three steps: we start from the frozen
StreamVLN model (Section~III-A), adapt its inference path for
resource-constrained execution (Section~III-B), and spend the recovered
resources on inline Stop Action verification (Section~III-C,
Fig.~\ref{fig:deployment}).

\subsection{Base Agent: StreamVLN}

Our system builds upon StreamVLN~\cite{streamvln}, a streaming VLN framework
with a LLaVA-style architecture: a SigLIP-SO400M-patch14-384 vision tower, a
two-layer MLP projector, and a Qwen2 language model ($28$ layers, $d=3584$),
$8.03$\,B parameters in total. StreamVLN combines a fast-streaming dialogue
context, held in a sliding-window KV cache, with a slow-updating memory
context, so inference never reprocesses the full navigation history. When the
window reaches its boundary the dialogue context is recomputed and repopulated
with stride-sampled memory tokens. 
Memory growth is bounded by a voxel-based
spatial token pruning rule, which we implement inside our VLN driver so that it runs on-board
(\S\ref{sec:quant}). It is inactive in the RGB-only VLN-CE setting, whose
evaluation agent supplies the encoder with a placeholder pose and a
zero-valued depth map.

\begin{figure}[!b]
  \centering
 \vspace{-0.5em}
 \includegraphics[width=1\linewidth]{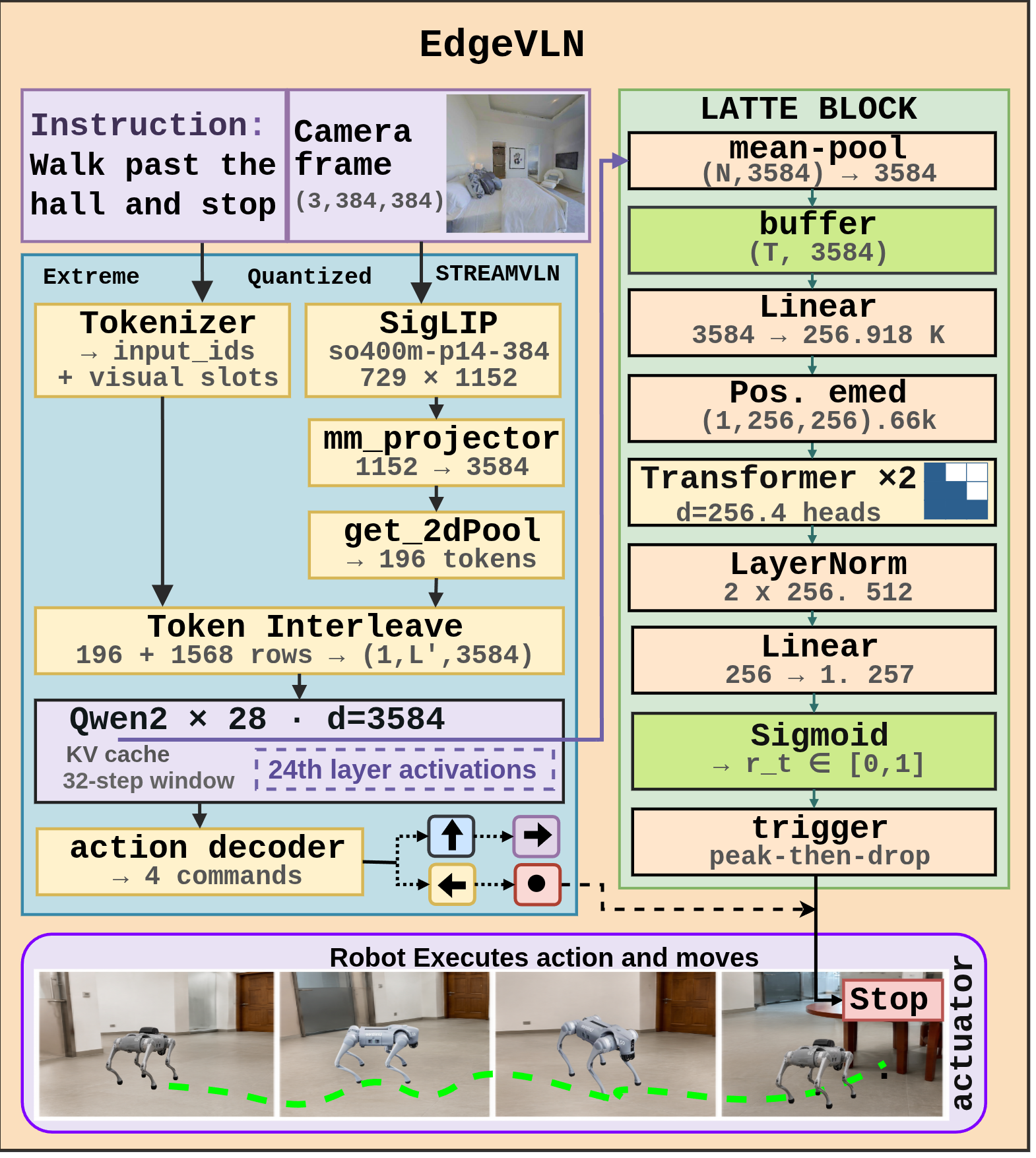}
  \caption{EdgeVLN inference path. \textbf{Input}: a language instruction and the current camera frame. The quantized StreamVLN model encodes them into a chunk of four actions, and LATTE reads the same forward pass's layer-24 activations to predict the Stop Action verifier rank. \textbf{Output}: the Stop Action verifier rank triggers \texttt{STOP} or the next action, which the robot executes to produce the next frame.}
  \label{fig:sys_arch}
\end{figure}

Both contributions below attach to the frozen model without retraining it, and are arranged as shown in Fig.~\ref{fig:sys_arch}.

\subsection{Quantization and Runtime Deployment Path
}
\label{sec:quant}


The board's 16\,GB must hold the operating system, runtime, KV cache, and compute buffers as well as the model, so full-precision weights cannot remain resident. We therefore quantize only the language-model feed-forward blocks, $5.70$\,B of $8.03$\,B parameters, leaving the vision tower, projector, attention projections, embeddings, and LM head at $16$ bits. We first sweep candidate precisions with fake quantization, which rounds weights to the target grid but stores them in $16$-bit form. This isolates precision from runtime: it measures each format's success rate in a controlled stack, so any later on-board gap is attributable to the deployment path rather than the format itself. The memory saving appears only in a runtime with native low-bit kernels. We therefore deploy with \texttt{llama.cpp}, which encodes one image per turn but provides no interface for the model's rolling context. Our driver reconstructs that context by exporting the vision tower and projector to the runtime's multimodal format, re-implementing the $27\!\times\!27\!\rightarrow\!14\!\times\!14$ pooling that yields $196$ tokens per frame, and clearing the context every $32$ steps while re-injecting stride-sampled earlier frames as raw embedding batches. The deployed stack thus consists of GGUF weights, our driver, and the runtime, with no deep-learning framework in the inference path.

Voxel-based spatial pruning is described in the original  StreamVLN work~\cite{streamvln} but is not released in its implementation, so we implement it directly in our \texttt{llama.cpp}-based driver. Each token cell takes the median of its valid depth pixels, is back-projected to a voxel index using the platform's depth and pose streams, and only the most recent token per voxel is retained within each temporal bucket. The pruning is applied only when visual history is re-injected, adding no latency on ordinary steps and incurring its cost at chunk boundaries, where the re-injected history is largest.


\subsection{LATTE: Latent Trajectory Termination Extractor }

The LATTE block reads the backbone's own layer-24 hidden state then mean-pools the 196 image-token rows of each frame to obtain a 3584-dimensional feature. These activations are already produced by the navigation forward pass, so the verifier adds neither a second vision encoder nor any extra backbone compute; only the head itself is new.
Decision features are accumulated in a buffer of up to $T=256$ decision steps and temporally averaged over $T=16$ steps before projection to $d=256$.

Because the supervision target is not a per-frame quantity, whether the agent has already passed its closest approach is a property of the trajectory, so the head is sequential rather than pointwise. Each buffered decision embedding is projected to $d=256$, combined with a learned positional embedding, and processed by two pre-norm transformer encoder layers with four heads and a feed-forward width of $512$. Attention is masked strictly causally with an unbounded history. A final LayerNorm and linear projection produce a scalar, followed by a sigmoid to obtain the Stop Action verifier rank $r_t\in[0,1]$ for each decision step. The head regresses \texttt{goal\_rank\_camera\_height} $=\mathrm{clip}\!\left(1-d_t/d_{\max},0,1\right)$ with $d_{\max}=5.0$\,m, zeroed after the closest-approach step and whenever the goal fails the camera-height visibility test, under a mean-squared-error loss.

The head is trained separately from the frozen model on 3184 collected successful rollouts features from R2R VLN-CE Dataset Train split, then split into 2548/318/318 train/val/test respectively at the episode level.

The predicted Stop Action verifier rank is consumed by a causal trigger with hysteresis. The
trigger tracks the running maximum of the score and issues \texttt{STOP} once
that maximum has exceeded a threshold and the current score has fallen a fixed
margin below it, so it reacts to a peak that has turned over rather than to
every local maximum. A minimum step count suppresses spurious early fires. Because incorrect stop decisions lower task success directly, the head is kept in FP32 and decoupled from the quantized backbone.

\section{Experiments}
\label{sec:exp}

\begin{figure}[t]
  \includegraphics[width=1\linewidth]{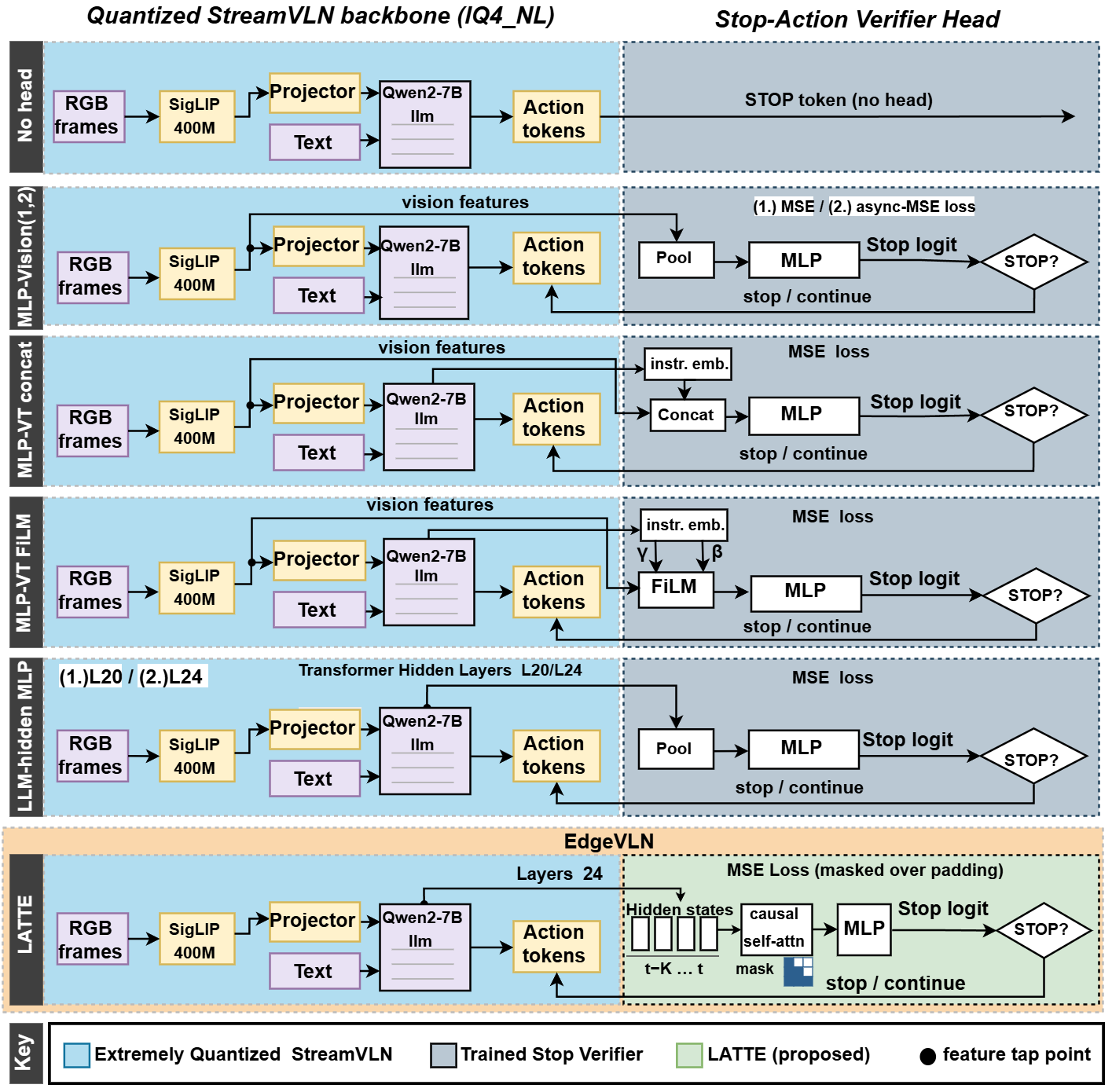}
  \caption{Overview of seven Stop Action verifier candidates evaluated on the R2R VLN-CE dataset validation-unseen split. All variants use a frozen StreamVLN backbone quantized to IQ4\_NL and different stop-prediction strategy: MLP-Vision(1,2), MLP-VT Concat, MLP-VT FiLM, LLM-hidden-MLP(L20,L24) and LATTE (ours), the selected candidate, taps intermediate Qwen2-7B hidden states and applies causal self-attention over recent hidden states for stop prediction. Gray denotes the frozen quantized backbone, white trained components, and green LATTE modules. The stop decision determines whether action generation continues, with MSE/asymmetric-MSE losses used by the corresponding heads.}
  \label{fig:stop-decision architecture}
  \vspace{-1.5em}
\end{figure}

In this section, we measure six precisions of the StreamVLN model and seven Stop Action verifier candidates~(Fig.~\ref{fig:stop-decision architecture}) on the two deployable backbones. We first evaluate SR and OSR in closed-loop simulation;
second, we measure latency, power, energy, and resident memory on-board by
replaying a recorded sequence of 89 navigation frames. The results are shown in Fig.~\ref{fig:onboard_cost} and Table~\ref{tab:r2r_results}.

\subsection{Experimental Setup}

Each configuration's accuracy is scored in Habitat simulator on the R2R VLN-CE Dataset val-unseen split \cite{vlnce}, over all $1839$ episodes, in closed loop: every action is executed and the next observation is rendered from the resulting pose, so a single incorrect action changes the remainder of the trajectory. We report success rate (SR), the fraction of episodes in which the agent itself issues
\texttt{STOP} within the success radius of the goal, and oracle success rate (OSR), the fraction whose executed trajectory passes within that radius at any point.
All configurations are scored on the same GPU, since identical models diverge
on different accelerators once a rollout is closed-loop. The Stop Action verification head's trigger threshold and drop margin are selected per backbone by a $16$-cell grid search on the same val-unseen episodes we report, formed by the Cartesian product of $\text{threshold}_\text{high} \in \{0.4,0.5,0.6,0.7\}$ and $\text{drop}_\text{margin} \in \{0.1,0.15,0.2,0.25\}$. Absent a held-out split, its gain should therefore be read as an upper bound.

\newcommand{\panttl}[2]{\par\vspace{2pt}{\footnotesize\textbf{(#1)}\quad #2\par\vspace{1pt}}}

\begin{figure}[!b]
  \vspace{-2em}
  \centering
  \panttl{a}{On-board measurement setup on Jetson Orin NX}
  \includegraphics[width=\linewidth]{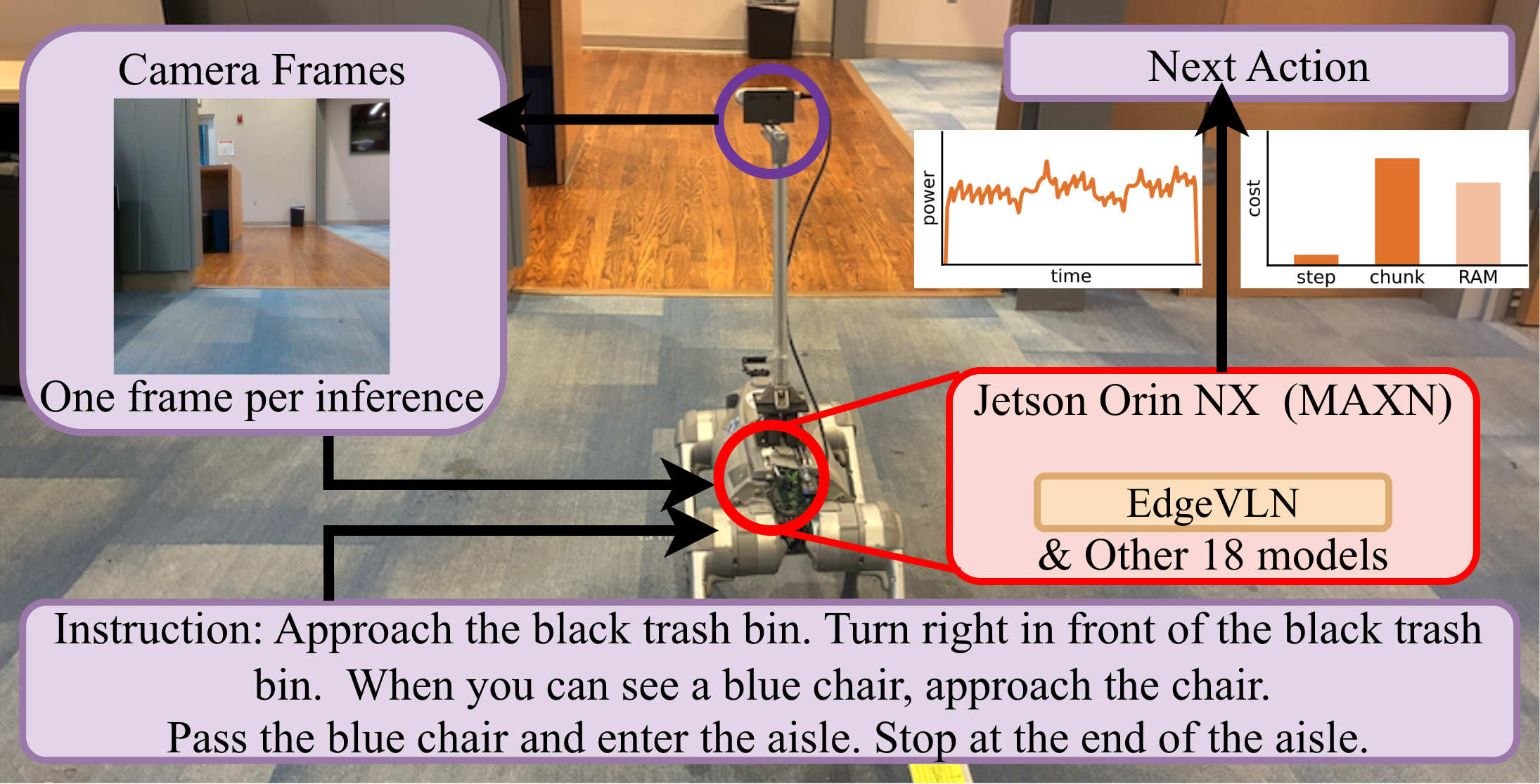}
  \panttl{b}{Module power over time for one 89-step episode}
  \includegraphics[width=\linewidth]{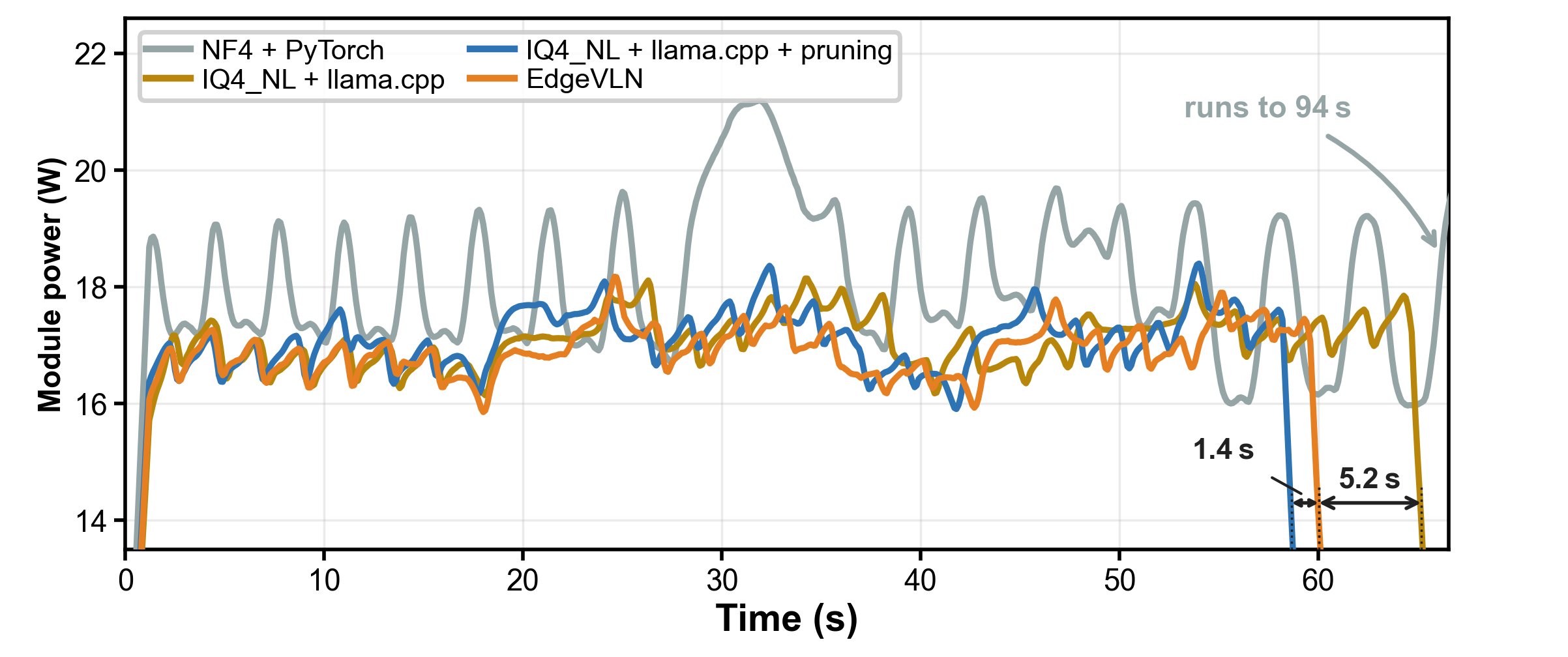}
  \panttl{c}{Step latency and peak memory}
  \includegraphics[width=\linewidth]{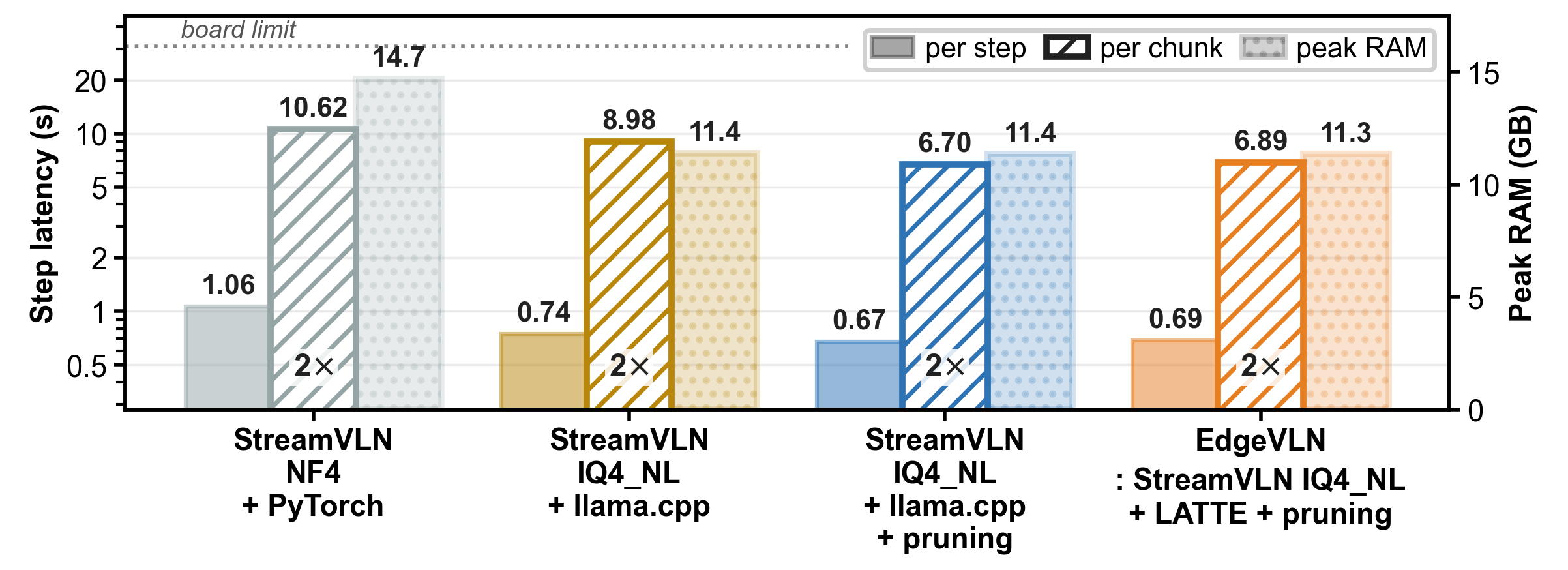}
  \caption{On-board measurement on the Jetson Orin NX. (a) Measurement setup in MAXN power mode: the robot stands still while the frames of one recorded real-world episode (89 frames) are replayed to it, so latency, power, energy, and memory are measured without a trajectory that changes from one configuration to another. (b) Both \texttt{llama.cpp} configurations draw less power and finish sooner than NF4 under PyTorch. (c) Mean step latency per configuration with peak resident memory; the number inside each per-chunk bar is how many chunk compressions occur in one episode. EdgeVLN improves success rate while nearly matching the quantized StreamVLN's on-board latency, energy, and memory under our driver.}
  \label{fig:onboard_cost}
\end{figure}

\begin{table}[t]
  \caption{Accuracy and on-board measurement results. (a) reports the six precisions of the StreamVLN model, from BF16 down to 2 bits; (b) reports seven stop action verifier candidates  with baseline, each on the two deployable backbones, BF16 and IQ4\_NL. SR and OSR come from $1{,}839$ closed-loop R2R VLN-CE Dataset val-unseen split episodes evaluation. We measured RAM, power, energy and latency on an NVIDIA Jetson Orin NX. As BF16 does not fit, only 20 of the 28 transformer blocks plus the output layer stay resident, the rest streamed from storage at each token.}
  \label{tab:r2r_results}
  \centering
  \scriptsize
  \setlength{\tabcolsep}{2pt}

  \textbf{(a)}~~StreamVLN under post-training quantization\par
  \vspace{2pt}
\begin{tabular}{clcccccc c}
\hline
\textbf{\#} & \kern1.5em\textbf{Quant.} & \textbf{Size} & \textbf{SR} $\uparrow$ & \textbf{OSR} $\uparrow$ & \textbf{RAM} $\downarrow$ & \textbf{Power} & \textbf{Energy} & \textbf{Step Latency} \\
 & & \textbf{(GB)} & \textbf{(\%)} & \textbf{(\%)} & \textbf{(GB)} & \textbf{(W)} & \textbf{(J)} & \textbf{(s)} \\
\hline
1 & \kern1.5emBF16 \cite{streamvln} & 16.06 & \textbf{57.69} & \textbf{65.04} & 14.99 & 10.85 & 151.68 & 13.98 \\
2 & \kern1.5emINT8 & 10.36 & 57.04 & 64.82 & 15.98 & \multicolumn{3}{c}{\textcolor{gray}{loads, but OOM after 8 steps}} \\
3 & \kern1.5emINT4 & 7.68 & 57.48 & 64.60 & 15.85 & 19.79 & 419.70 & 21.21 \\
4 & \kern1.5emNF4 & 7.68 & 57.15 & 64.76 & 14.70 & 18.18 & 19.31 & 1.06 \\
5 & \kern1.5em\llap{\kern0.0em\textcolor{green!55!black}{\ooalign{$\checkmark$\cr\kern0.07em$\checkmark$\cr}}\kern0.25em}IQ4\_NL & 7.94 & 57.31 & 64.11 & \textbf{11.35} & 16.97 & \textbf{11.41} & \textbf{0.67} \\
6 & \kern1.5emINT2 & 6.44 & 46.00 & 55.79 & 16.03 & 20.08 & 437.86 & 21.81 \\
\hline
\end{tabular}

  \vspace{0.8em}
  \textbf{(b)}~~ Results of seven Stop Action verifier candidates\par
  \vspace{2pt}
\begin{tabular}{lccccc c}
\hline
\textbf{Base model} & \textbf{SR} $\uparrow$ & \textbf{OSR} $\uparrow$ & \textbf{RAM} $\downarrow$ & \textbf{Power} & \textbf{Energy} & \textbf{Step Latency} \\
 & \textbf{(\%)} & \textbf{(\%)} & \textbf{(GB)} & \textbf{(W)} & \textbf{(J)} & \textbf{(s)} \\
\hline
\multicolumn{7}{l}{\textit{no head}} \\
\hspace{2ex}BF16 \cite{streamvln} & 57.69 & \textbf{65.04} & 14.99 & 10.85 & 151.68 & 13.98 \\
\hspace{2ex}IQ4\_NL & 57.31 & 64.11 & 11.35 & 16.97 & \textbf{11.41} & \textbf{0.67} \\
\hline
\multicolumn{7}{l}{\textit{MLP vision (MSE) (+\,5.8\,MB)}} \\
\hspace{2ex}BF16 & 51.70 & 57.50 & 14.76 & 10.92 & 150.01 & 13.74 \\
\hspace{2ex}IQ4\_NL & 51.50 & 56.70 & 11.35 & 16.97 & 11.42 & \textbf{0.67} \\
\cline{1-7}
\multicolumn{7}{l}{\textit{MLP vision (asym MSE) (+\,5.8\,MB)}} \\
\hspace{2ex}BF16 & 55.00 & 62.20 & 14.73 & 10.92 & 150.32 & 13.77 \\
\hspace{2ex}IQ4\_NL & 54.80 & 61.30 & \textbf{11.34} & 16.94 & 11.44 & 0.68 \\
\cline{1-7}
\multicolumn{7}{l}{\textit{MLP vision+text concat (+\,20.5\,MB)}} \\
\hspace{2ex}BF16 & 45.79 & 50.68 & 14.73 & 10.91 & 150.35 & 13.79 \\
\hspace{2ex}IQ4\_NL & 45.46 & 49.76 & 11.40 & 16.90 & 11.49 & 0.68 \\
\cline{1-7}
\multicolumn{7}{l}{\textit{MLP vision+text FiLM (+\,38.8\,MB)}} \\
\hspace{2ex}BF16 & 55.14 & 61.39 & 15.00 & 10.93 & 147.60 & 13.50 \\
\hspace{2ex}IQ4\_NL & 54.70 & 60.47 & 11.43 & 16.87 & 11.51 & 0.68 \\
\cline{1-7}
\multicolumn{7}{l}{\textit{LLM hidden L20 (+\,15.7\,MB)}} \\
\hspace{2ex}BF16 & 58.24 & 64.17 & 14.69 & 10.89 & 150.53 & 13.83 \\
\hspace{2ex}IQ4\_NL & 57.86 & 63.84 & 11.40 & 16.86 & 11.43 & 0.68 \\
\cline{1-7}
\multicolumn{7}{l}{\textit{LLM hidden L24 (+\,15.7\,MB)}} \\
\hspace{2ex}BF16 & 58.13 & 64.60 & 14.65 & 10.91 & 155.11 & 14.22 \\
\hspace{2ex}IQ4\_NL & 57.86 & 63.73 & 11.37 & 16.93 & 11.47 & 0.68 \\
\cline{1-7}
\multicolumn{7}{l}{\textbf{LATTE:}~\textit{LLM hidden L24 + causal self-attn (+\,8.2\,MB)}} \\
\hspace{2ex}BF16 & \textbf{58.35} & 64.44 & 14.65 & 10.90 & 148.91 & 13.67 \\
\hspace{2ex}\llap{\textcolor{green!55!black}{\ooalign{$\checkmark$\cr\kern0.07em$\checkmark$\cr}}\,}IQ4\_NL & 58.02 & 63.46 & 11.35 & 16.84 & 11.54 & 0.69 \\
\hline
\end{tabular}

  \vspace{-2em}
\end{table}

These measurements are taken on the Jetson Orin NX board shown in Fig.~\ref{fig:deployment}, with the robot held stationary throughout. A closed-loop rollout would take a different trajectory for every configuration and confound the measurements with episode length, so we instead recorded one real-world driving episode with the robot's camera and replay its 89 frames to
the board, giving every configuration a byte-identical observation sequence. Latency, power, energy and resident
memory are measured on an NVIDIA Jetson Orin NX 16\,GB (JetPack R35.3.1),
whose unified pool is $16.14$\,GB, in the MAXN power mode. Power is measured
on the Jetson module's supply rail, so it covers the board alone and not the
robot's motors, and the board's idle draw is not subtracted. 
Energy and latency per navigation step are the episode totals divided by its 89 steps;
one generation emits a
four-action chunk, so a per-decision figure is the per-step figure times
$89/24$. Each configuration is repeated three times under matched thermal
conditions and we report the mean; peak resident memory is taken from the
repeat whose starting residency matched idle, because page cache carried over
from a previous run inflates it. Within a backbone the measured differences
between heads fall below this run-to-run spread and should not be compared;
the BF16 rows are single runs.

\subsection{Experimental Results}
Quantization halves the deployed model without degrading navigation quality.
Table~\ref{tab:r2r_results}(a) shows the model
shrinking from $16.06$ to $7.94$\,GB, while SR moves from $57.69\%$ to
$57.31\%$ and OSR from $65.04\%$ to $64.11\%$. Every 4-bit format stays within
$0.54$ SR points of full precision, so the choice among four-bit formats is
settled by latency and energy rather than success rate. Only 2 bits breaks the model, at
$46.00\%$ SR.

The efficiency this buys depends on the execution path rather than on the bit width alone. INT4 and NF4 hold $7.68$\,GB of weights, less than the $7.94$\,GB
of IQ4\_NL, yet peak at $15.85$ and $14.70$\,GB resident under PyTorch against
$11.35$\,GB under our VLN driver; INT8 at $10.36$\,GB
loads but is killed after eight navigation steps. The same reversal appears in
energy, and far more sharply (Fig.~\ref{fig:onboard_cost}, Table~\ref{tab:r2r_results}(a)): $11.41$\,J per
step for IQ4\_NL against $19.31$\,J for NF4 ($1.7\times$) and $419.70$\,J for
INT4 ($36.8\times$), whose dequantization path leaves each step at $21.21$\,s.
Full precision is the extreme case: BF16 runs with only 20 of its 28 transformer blocks plus the output layer resident
at $20.8\times$ the latency and $13.3\times$ the energy of IQ4\_NL for $0.38$ SR points more, while drawing the lowest
average power in the table ($10.85$\,W) because the GPU idles waiting on storage.

At the deployed operating point one navigation step takes $0.672$\,s and
$11.41$\,J at $16.97$\,W sustained ($2.49$\,s and $42.3$\,J per four-action
decision), since only $24$ of the $89$ requests trigger a generation and the
rest return an already buffered action in $0.024$\,s. Voxel pruning removes
$41.7\%$ of the re-injected memory tokens across the $3{,}252$ chunk
boundaries of the full run, which is what brings the amortized step from
$0.744$ to $0.672$\,s, at $+0.05$ points of success rate, one episode of $1839$.

The memory savings are enough to recover the lost success rate, while Table~\ref{tab:r2r_results}(b) shows that what the new Stop Action verification head reads determines whether recovery is possible.
Heads over pooled vision features lose accuracy outright, from
$57.69\%$ to $51.70\%$ on BF16, and fusing the instruction does not rescue
them: concatenation is the worst head at $45.79\%$, and FiLM ($55.14\%$) merely
returns to the vision-only level. Only heads over the decoder's own hidden
states gain, reaching $58.24\%$ at layer 20 and $58.13\%$ at layer 24. LATTE
reads the same layer-24 states through causal self-attention rather than a
pointwise MLP and improves on both at half the size, $58.35\%$ on BF16 and
$58.02\%$ on IQ4\_NL at $8.2$\,MB against $15.7$\,MB. The two backbones track
each other to within $0.33$ points, comparable to the 0.38 points without the head, so
the gain carries over to the quantized model. On IQ4\_NL it corrects $26$
episodes and breaks $13$, and it acts where the design predicts: on
termination rather than on path-following, raising SR by $0.71$ points while lowering OSR by $0.65$ points.

LATTE leaves deployed efficiency essentially unchanged. It adds $0.013$\,s and $0.13$\,J per navigation step ($0.048$\,s and $0.48$\,J per decision, or $+1.9\%$ and $+1.1\%$) without an additional forward pass, since it consumes activations already produced by navigation. Average power falls slightly, from $16.97$ to $16.84$\,W, because the head runs on the CPU and adds a phase in which the GPU is idle; as in BF16, energy is therefore the more meaningful metric.

\section{Conclusion}

We introduce EdgeVLN, a deployment-ready quantized vision--language navigation model combining StreamVLN and LATTE, an inline Stop Action verifier, executed by our \texttt{llama.cpp} VLN driver on a Jetson Orin NX 16,GB. Across 1,839 closed-loop R2R VLN-CE val-unseen episodes, all four-bit formats achieve nearly identical success rates, yet their on-board step energy spans $36.8\times$ by execution path. Only IQ4\_NL under our driver fits, with 11.35,GB resident memory, $20.8\times$ higher speed, and $13.3\times$ lower step energy than storage-streamed BF16. Thus, runtime—not bit width—determines whether the model fits. LATTE reuses hidden activations from the frozen backbone without an extra forward pass, raising the deployed four-bit model to 58.02\% SR, above BF16, while adding under 2\% of the step's latency and energy.



\bibliographystyle{IEEEtran}
\bibliography{references,eehpc}

\section*{Author Biographies}

\noindent
\textbf{Rithvik Jonna} is a Ph.D. student in Electrical and Computer Engineering at Johns Hopkins University, Whiting School of Engineering. His research interests include embedded AI systems, energy-efficient machine learning, robotics, with a focus on deploying multimodal perception systems on resource-constrained robotic platforms. Contact him at djonna1@jh.edu.

\medskip

\noindent
\textbf{Man Namgung} is an M.S.E. student in Robotics at Johns Hopkins University, Baltimore, where he is a graduate researcher in the Energy Efficient High Performance Computing (EEHPC) Lab. His research interests include embodied multimodal AI, autonomous systems, and deep learning. Namgung received B.S. degrees in computer science and in military arts and science from the Republic of Korea Naval Academy, Changwon, South Korea. Contact him at mnamgun1@jh.edu.

\medskip

\noindent
\textbf{Aakash Gurram} is pursuing the M.S.E. degree in Robotics at Johns Hopkins University, Baltimore, where he is also a Graduate Research Assistant with the Energy-Efficient High-Performance Computing (EEHPC) Laboratory. He received the B.S. degree in Robotics and Automation from Karunya Institute of Technology and Sciences, Coimbatore, India, in 2024. His research interests include embodied AI, autonomous navigation, and machine learning. Contact him at agurram1@jh.edu.

\medskip

\noindent
\textbf{Tinoosh Mohsenin} is an Associate Professor of Electrical and Computer Engineering at Johns Hopkins University, Whiting School of Engineering. Her research focuses on energy-efficient computing for signal processing, machine learning, and multimodal AI in applications including autonomous systems, health monitoring, and cyber-physical systems. She is affiliated with the Laboratory for Computational Sensing and Robotics (LCSR), and directs the Energy-Efficient High-Performance Computing Lab (EEHPC). Contact her at tinoosh@jhu.edu.

\end{document}